\documentclass{article}
\usepackage{ijcai15}
\usepackage{times}
\usepackage{amsmath}
\usepackage{amsfonts,amssymb}
\usepackage{bm}
\usepackage{amsthm}
\usepackage{graphicx}
\usepackage{url}
\usepackage[english]{babel}
\usepackage{subfigure}
\usepackage{caption}
\usepackage{algorithm} 
\usepackage{algorithmic} 

\theoremstyle{definition}
\newtheorem{definition}{Definition}[section]

\theoremstyle{remark}

\title{Action2Activity: Recognizing Complex Activities from Sensor Data}
\author{Ye Liu$^\dag$, Liqiang Nie$^\dag$, Lei Han$^{\S}$, Luming Zhang$^\dag$, David S. Rosenblum$^\dag$\\
        {$^{\dag}$ School of Computing, National University of Singapore}\\
        {$^{\S}$ Department of Computer Science, Hong Kong Baptist University}\\
        \{liuye, david\}@comp.nus.edu.sg, leihan@comp.hkbu.edu.hk, \{nieliqiang, zglumg\}@gmail.com\\
        }
\begin{document}
\maketitle
\begin{abstract}
As compared to simple actions, activities are much more complex, but semantically consistent with a human's real life. Techniques for action recognition from sensor generated data are mature. However, there has been relatively little work on bridging the gap between actions and activities. To this end, this paper presents a novel approach for complex activity recognition comprising of two components. The first component is temporal pattern mining, which provides a mid-level feature representation for activities, encodes temporal relatedness among actions, and captures the intrinsic properties of activities. The second component is adaptive Multi-Task Learning, which captures relatedness among activities and selects discriminant features. Extensive experiments on a real-world dataset demonstrate the effectiveness of our work.
\end{abstract}

\section{Introduction}\label{sec:intro}
We are living in an era of wearable and environmental sensors. Activity recognition from sensor data plays an essential role in many applications. Consider application scenarios in healthcare as an example~\cite{Liqiang2015}. Caregivers use sensors to track and analyze the Activities of Daily Living (ADL) of elderly people, which enables the caregivers to provide proactive assistance~\cite{PansiotSensorFusion2007}. Another application scenario is context-aware music recommendation~\cite{wang2012context}, which senses context information about the activity a user is doing and recommends music suitable for the activity.

Several research efforts have been dedicated to the recognition of simple activities\footnote{\small{In this work, we use the term activity to stand for a complex activity, and we use the term action to refer to a simple activity.}} with simple features~\cite{wang2012context,PansiotSensorFusion2007,ravi2005activity}. For example, Ravi \emph{et al.}~\shortcite{ravi2005activity} designed a classifier to distinguish eight actions, namely \emph{standing, walking, running, climbing up stairs, climbing down stairs, vacuuming, brushing teeth, and sit-ups}. In real life, however, human activity is much more complex than such disjoint occurrences of simple actions. Consider cooking as an example. It involves a sequence of actions over time, and some of those actions may happen simultaneously or concurrently, such as walking followed by reaching for the fridge, or standing while fetching or cutting up food.

Recognizing complex activities from sensor data is non-trivial due to the following reasons. First, the actions underlying an activity are not independent. In particular, within one activity, the temporal relatedness among actions may manifest itself in many forms. Such sophisticated temporal combinations lead to semantic meanings for activity understanding. Second, commonality or semantic relatedness exists across multiple activities. For example, making coffee may be much more similar to making bread than to relaxing, in terms of action patterns. Third, features used to represent activities usually suffer from the curse of dimensionality, and in fact not all features are discriminative.

To tackle the above challenges, we present an approach that consists of two components. The first is an algorithm for temporal pattern mining, which encodes various temporal relations among actions, including sequential, interleaved, and concurrent relations. In particular, we presume that each pattern is a set of temporally inter-related actions, and that each activity can be characterized by a set of patterns. Our algorithm automatically discovers frequent temporal patterns from action sequences and uses the mined patterns to characterize activities. The second component is activity recognition. It treats each activity as a task and uses an adaptive multi-task learning (\textbf{aMTL}) algorithm to capture and model the relatedness among these tasks. In addition, it is capable of identifying discriminant task-specific and task-sharing features. As an added benefit, it alleviates the problem of insufficient training samples, since it enables sharing of training instances among tasks.

We summarize the contributions as follows:
\begin{itemize}
  \item We present a novel approach to identify temporal patterns among actions for activity representation.
  \item We present a novel multi-task learning approach to boost the performance of activity recognition.
\end{itemize}

\section{Related Work}
Recognizing simple actions from sensor data has attracted much attention~\cite{wang2012context,liu2010visual,liu2012fusion,cui2013tracking,lu2016towards,PansiotSensorFusion2007,ravi2005activity}. For example, the work introduced by Ravi \emph{et al.}~\shortcite{ravi2005activity} classified eight daily actions using shallow classifiers, such as kNN, SVM and Na\"{\i}ve Bayes, and achieved overall accuracy of $95\%$. Another work by Wang \emph{et al.}~\shortcite{wang2012context} utilized a Na\"{\i}ve Bayes model to recognize six daily actions (\emph{working, studying, running, sleeping, walking and shopping}) for music recommendation and obtained promising performance. However, as the nature of human activity is complex, people often perform not just a single action in isolation, but several actions in diverse combinations. The key to modeling activities is to capture the temporal relations among actions~\cite{zhuo_IJCAI2009_learning}. Few of the previous efforts have been dedicated to capturing the relatedness among actions and the high-level semantics over groups of actions.

A set of approaches have been proposed to explore the simple relations among actions~\cite{yang_IJCAI2009_activity,liu2016action,ryoo2006recognition_CVPR}. Dynamical model approaches (e.g., HMM~\cite{HMM_rabiner1989tutorial} and CRF~\cite{CRF_lafferty2001conditional}) could capture the simple relations between actions, such as sequential relations. However, they are unable to characterize the complex relations in real-world activity data. This is because human activities may contain several overlapped actions, and treating these activities simply as sequential data may lead to information loss. Moreover, they fail to capture the higher-order temporal relatedness among actions. Bayesian network-based approaches~\cite{ITBN_zhang2013modeling} are also able to model the temporal relations among actions. Bayesian network models use the directed acyclic graph for both learning and inference, they hence face the problem of handling temporal relation conflicts among actions. Pattern-based approaches try to capture the complex temporal relatedness via temporal patterns and have demonstrated their advantages in handling the relatedness problem in the medical and finance domains~\cite{patel_TemporalMining_KDD2008,wu_TKDE2007_TemporalMining}. However, as far as we know, the literature on temporal pattern-based representations for sensor-based activity recognition is relatively sparse. Gu et al.~\shortcite{gu2009epsicar} proposed an Emerging Pattern (EP)-based approach for activity recognition. Their approach is able to handle sequential, interleaved and concurrent relations between pairwise actions. In contrast to their work, ours provides a more general way to describe temporal relations among more than two actions. Moreover, our method can capture the intrinsic relatedness among various activities, which can further enhance overall recognition performance.

Multi-task learning (MTL) is a learning paradigm that jointly learns multiple related tasks and can achieve better generalization performance than learning each task individually, especially with those insufficient training samples. The relations among tasks can be pairwise correlations~\cite{ZhangYu_UAI_MTL_2010}, or pairwise correlation within a group~\cite{zhou_NIPS2011_clusteredMTL}, as well as higher-order relationships ~\cite{zhang_IJCAI2013_MTLHighOrderLearning}. However, for activity recognition, encoding only task relatedness is not enough. Since not all features are discriminative for the prediction tasks, it is reasonable to assume that only a small set of features is predictive for specific tasks. In the light of this, group Lasso~\cite{GroupLassoOriginal_Statistic2006} is a technique used for selecting group variables that are key to the prediction tasks.  As an important extension of Lasso~\cite{tibshirani_LASSO1996regression}, group Lasso combines the feature strength over all tasks and tends to select the features based on their overall strength. It ensures that all tasks share a common set of features, while each one keeps its own specific features~\cite{ZhouJiayu_KDD2011_GroupLasso}.

\section{Temporal Pattern Mining}
A training collection for activity recognition consists of multiple activities, and each activity is a sequence of actions with its corresponding start-time and end-time. We aim to mine the frequent temporal patterns from these sequences and use the patterns to represent activities for subsequent learning. We define some important concepts first.

\begin{definition}
Let $\Theta$ denote the action space. An \textbf{\emph{action}} $act\in \Theta$ that occurs during a period of time is denoted as a triplet $ a=(t_{s}, id_{a}, t_{e})$, where $id_{a}$ is the action $id$, $t_{s}$ is the start-time, $t_{e}$ is the end-time, and $t_{s}<t_{e}$. Each \textbf{\emph{activity}} over $\Theta$ is defined as a sequence of actions ordered by start-time. Formally, $activity =\  < a^1, a^2, \ldots, a^p > $, with the constraints $a^i=(t^{i}_{s}, id^{i}_{a}, t^{i}_{e})$, and $t^{i}_{s}\leq t^{i+1}_{s}$ for $1\leq i\leq p-1$.
\end{definition}

\begin{definition}
A \emph{\textbf{temporal pattern}} $P$ of dimension $k>1$ is defined as a pair $(\mathcal{S}, \textbf{R})$, where $\mathcal{S} \subseteq \Theta$ is a set of actions. $\textbf{R}$ is a $k \times k$ matrix, where $R_{i,j}$, its $(i,j)$-$th$ element, indicates the temporal relatedness between $act_i \in \mathcal{S}$ and $act_j\in \mathcal{S}$.  The temporal relation is encoded with Allen's temporal interval logic \cite{allen1983maintaining}. The dimension of a temporal pattern $P$ is written as $dim(P)$, which equals $|\mathcal{S}|$. If $dim(P) = k$, then temporal pattern $P$ is called a \emph{$k$-pattern}.
\end{definition}

\begin{definition}
We denote the \textbf{\emph{partial order}} over temporal patterns as $\sqsubseteq$ and define it as follows: The temporal pattern $(\mathcal{S}_A, \textbf{R}_A)$ is a subpattern of temporal pattern $(\mathcal{S}_B, \textbf{R}_B)$ (or $(\mathcal{S}_A, \textbf{R}_A) \sqsubseteq (\mathcal{S}_B, \textbf{R}_B)$), if and only if it satisfies these two conditions: $(a)$ $dim((\mathcal{S}_A, \textbf{R}_A))<dim((\mathcal{S}_B, \textbf{R}_B))$; and $(b)$ there exists an injective mapping $\pi :\{1, \ldots, dim((\mathcal{S}_A, \textbf{R}_A))\}\rightarrow \{1, \ldots, dim(\mathcal{S}_B, \textbf{R}_B)\}$ such that $\forall i,j \in \{1, \ldots, dim((\mathcal{S}_A, \textbf{R}_A))\}: \mathcal{S}_A(i) = \mathcal{S}_B(\pi(i)) \wedge \textbf{R}_A[i,j] = \textbf{R}_B[\pi(i), \pi(j)]$. In particular, the equality of actions in condition (b) depends only on the action $id$ and is independent of the start and end times.
\end{definition}

\begin{definition}
The \textbf{\emph{support}} of a temporal pattern $P$ is defined as $sup(P) = \frac{L_P}{L}$, where $L_P$ refers to the total time that the pattern
can be observed within a sliding window. $L = t_{win} + t_{activity}$ is a normalizing term, where $t_{win}$ denotes the width of the sliding window and $t_{activity}$ refers to the total time of the activity \cite{hoppner2001discovery}. In practice, the window length can be set as the maximum or average length of actions in the dataset. We can interpret the support $sup(P)$ as the observation probability of pattern $P$ within the given activity. This gives us the intuition of choosing this support for describing action patterns: the larger the support value for a given pattern is, the more correlated the pattern will be with a complex activity.
\end{definition}

\begin{definition}
Denote the minimum support threshold as $minsup$. Then a pattern $P$ is regarded as a \textbf{\emph{frequent pattern}} if $sup(P)\geq minsup$.
\end{definition}

Basically, the notion of frequent pattern is utilized to bridge the semantic gap between actions and activities. It captures the intrinsic
descriptions of activities and hence provides a natural way to informatively represent activities for further recognition. To discover the discriminant temporal patterns, we initially estimate the support value of each action (which is the simplest pattern, i.e., a $1$-$pattern$). After that, we remove the $1$-$patterns$ with small support values. Based upon the remaining $1$-$patterns$, we generate the $2$-$patterns$ and calculate their support values. Similarly, we prune the infrequent $2$-$patterns$. By iterating this procedure $k$ times, we ultimately obtain a set of patterns with dimensions up to $k$. The algorithm terminates when no more frequent patterns are found. The pruning process is effective due to the \textbf{\emph{Apriori Principle}}
and \textbf{\emph{Monotonicity Property}}~\cite{AgrawalApriori1994}. Notedly, the support of a pattern is always less than or equal to the support of any of its sub-patterns. In other words,
\begin{equation*}\label{suboptimility}
  \forall patterns \  P, Q: Q \sqsubseteq P \Rightarrow sup(Q)\geq sup(P).
\end{equation*}
This property is guaranteed by the definition of partial order and the support value of temporal patterns~\cite{hoppner2001discovery}. With the above property, our algorithm does not miss any frequent pattern. Moreover, this property ensures the correctness of our pruning step since all the sub-patterns of a frequent temporal pattern must also be frequent.

With the mined temporal patterns, we put all of them together to construct a joint pattern feature space. Thereby, each activity can be represented within this feature space, and each entry in the feature vector is the support value of the corresponding pattern. According to the definition, the support of a temporal pattern is conditioned on a specific activity, and the relevance of the support with respect to a specific activity is implied in the support estimation step. Thus, it is intuitive to describe activities in this way, since the higher the support value for a given pattern, the more relevant or important the pattern is for the given activity. It is worth mentioning that the feature space is generated from the training data only, and both the testing and training data share the same pattern feature space.

\section{Adaptive Multi-Task Learning}
Similar activities may share some patterns. For example, ``playing badminton" shares many similar temporal patterns with ``playing tennis", but greatly differs from ``fishing". Moreover, the dimension of temporal pattern features is usually very high, but not all temporal patterns are sufficiently discriminative for activity recognition. To capture the relatedness among different activities and select the discriminative patterns simultaneously, we regard each activity recognition as a task and present an \textbf{aMTL} model. It is able to adaptively capture the relatedness among tasks, as well as learn the task-sharing and task-specific features.

\subsection{Problem Formulation}
We first define some notations. In particular, we use bold capital letters (e.g., \textbf{X}) and bold lowercase letters (e.g., \textbf{x}) to denote matrices and vectors, respectively. We employ non-bold letters (e.g., x) to represent scalars, and Greek letters (e.g., $\lambda$) as parameters. Unless stated, otherwise, all vectors are in column form. Assume that we have $M$ kinds of activities/tasks \{$T_{1}, T_{2}, ..., T_{M}$\} in the given training set $\Phi$. $\Phi$ is composed of $N$ samples \{$(\textbf{x}_{1}, \textbf{y}_{1}),(\textbf{x}_{2}, \textbf{y}_{2}),...,(\textbf{x}_{N}, \textbf{y}_{N})$\}. Each training sample is a sequence of actions, represented by temporal pattern-based feature vector $\textbf{x}_{i} \in \mathbb{R}^{D}$ and their corresponding label vector $\textbf{y}_{i} \in \mathbb{R}^{M}$, where $\textbf{y}_{i}$ is the label vector with a single one and all other entries zero. Each instance is only one activity, and $D$ is the feature dimension. The prediction model for task $T_{i}$ of a given sample is defined as $f_{i}(\textbf{x}) = \textbf{x}^{T}\textbf{w}^{i}$, where $\textbf{w}^{i}\in \mathbb{R}^{D}$ is the weight vector for task $T_{i}$. Let $\textbf{X} = [\textbf{x}_{1}, \textbf{x}_{2}, \ldots, \textbf{x}_{N}]^{T}\in \mathbb{R}^{N\times D}$ be the data matrix and $\textbf{Y} = [\textbf{y}_{1}, \textbf{y}_{2}, \ldots, \textbf{y}_{N}]^{T}\in \mathbb{R}^{N\times M}$ be the label matrix. The weight matrix over $M$ tasks is denoted as $\textbf{W} = [\textbf{w}^{1}, \textbf{w}^{2}, \ldots, \textbf{w}^{M}]\in \mathbb{R}^{D\times M}$.

We formulate activity recognition as,
\begin{small}
\begin{eqnarray}\label{ObjectiveFunction}
  \nonumber
  \mathop{\min}_{\textbf{W},\bm{\Omega}} & \frac{1}{2} \| \textbf{X}\textbf{W} - \textbf{Y} \| ^{2}_{F} + \lambda tr(\textbf{W} \bm{\Omega}^{-1} \textbf{W}^{T}) + \gamma\| \textbf{W}\| ^{2}_{F} + \theta\| \textbf{W} \|_{2,1}, & \\
  & s.t. \quad \bm{\Omega} \succeq 0, \quad tr(\bm{\Omega}) = 1, &
\end{eqnarray}
\end{small}where the first term measures the empirical error; the second term adaptively encodes the relatedness among different tasks; the third term controls the generalization error; and the last one is the group Lasso penalty which helps to select the desired features automatically. $\lambda, \gamma, \theta$ are the regularization parameters, and $\bm{\Omega}\in \mathbb{R}^{M\times M}$ is a positive semi-definite matrix that we aim to learn. The $(i,j)$-th entry in $\bm{\Omega}$ represents the relations between task $i$ and task $j$. As an improvement over a uniform or pre-defined relatedness model~\cite{kato_NIPS2008_MTL}, we adaptively learn the relatedness among tasks. The $\ell_{2,1}$-norm of a matrix $\textbf{W}$ is defined as $\| \textbf{W} \|_{2,1}=\sum_{i=1}^{d}\sqrt{\sum_{j=1}^{M}W_{ij}^2}$. In particular,  $\ell_{2,1}$-norm applies an $\ell_{2}$-norm to each row of $\textbf{W}$ and these $\ell_{2}$-norms are combined through an $\ell_{1}$-norm. Thus, the weights of one feature over $M$ tasks are combined through $\ell_{2}$-norm and all features are further grouped via $\ell_{1}$-norm. The $\ell_{2,1}$-norm thereby plays the role of selecting features based on their strength over all tasks. As we assume that only a small set of features are predictive for each recognition task, the group Lasso penalty ensures that all activities share a common set of features while still keeping their activity-specific features.

\subsection{Optimization}
Eqn.$(\ref{ObjectiveFunction})$ is convex with respect to $\textbf{W}$ and $\bm{\Omega}$. We adopt the alternative optimization procedure to solve it.

\textbf{Optimizing $\textbf{W}$ with $\bm{\Omega}$ fixed}.
When $\bm{\Omega}$ is fixed, the optimization problem in Eqn.$(\ref{ObjectiveFunction})$ becomes an unconstrained convex optimization problem. The optimization problem can be rewritten as follows,
\begin{small}
\begin{eqnarray}\label{OptimizingW}
  \mathop{\min}_{\textbf{W}} & \frac{1}{2} \| \textbf{XW} - \textbf{Y} \| ^{2}_{F} + \lambda tr(\textbf{W} \bm{\Omega}^{-1} \textbf{W}^{T}) + \gamma \| \textbf{W} \| ^{2}_{F} + {\theta}\| \textbf{W} \|_{2,1}.&
\end{eqnarray}
\end{small}Since the objective function in Eqn.$(\ref{OptimizingW})$ is convex and non-smooth, we use the Fast Iterative Shrinkage-Thresholding Algorithm (FISTA)~\cite{beck_SIAM2009_FASTA} to solve it.

\textbf{Optimizing $\bm{\Omega}$ with $\textbf{W}$ fixed}.
When $\textbf{W}$ is fixed, the optimization problem in Eqn.$(\ref{ObjectiveFunction})$ for solving $\bm{\Omega}$ becomes
\begin{small}
\begin{eqnarray}\label{OptimizingOmega}
      \nonumber 
      \mathop{\min}_{\bm{\Omega}}  & tr(\bm{\Omega}^{-1} \textbf{W}^{T}\textbf{W}),& \\
      s.t.  & \bm{\Omega}\succeq 0, \quad tr(\bm{\Omega}) = 1. &
\end{eqnarray}
\end{small}
Denote $\textbf{A} =\textbf{W}^{T}\textbf{{W}}$
\begin{small}
\begin{eqnarray}
  \nonumber
  tr(\bm{\Omega}^{-1}\textbf{A}) &=& tr(\bm{\Omega}^{-1}\textbf{A})tr(\bm{\Omega}) \\
  \nonumber
  &=& tr((\bm{\Omega}^{-\frac{1}{2}}\textbf{A}^{\frac{1}{2}})(\textbf{A}^{\frac{1}{2}}\bm{\Omega}^{-\frac{1}{2}}))tr(\bm{\Omega}^{\frac{1}{2}}\bm{\Omega}^{\frac{1}{2}}) \\
  &\geq& (tr(\bm{\Omega}^{-\frac{1}{2}}\textbf{A}^{\frac{1}{2}}\bm{\Omega}^{\frac{1}{2}}))^{2} = (tr(\textbf{A}^{\frac{1}{2}}))^{2}.
\end{eqnarray}
\end{small}The equality holds if and only if $\bm{\Omega}^{-\frac{1}{2}}\textbf{A}^{\frac{1}{2}}=a\bm{\Omega}^{\frac{1}{2}}$ for some constant $a$ and $tr(\bm{\Omega}) = 1$. Therefore, there exists a closed-form solution for $\bm{\Omega}$, i.e.,
\begin{small}
\begin{equation}\label{OmegaAnalyticaSolution}
  \bm{\Omega} = \frac{(\textbf{W}^{T}\textbf{W})^{\frac{1}{2}}}{tr((\textbf{W}^{T}\textbf{W})^{\frac{1}{2}})}.
\end{equation}
\end{small}

\section{Experiments}

\subsection{Dataset}
The \emph{Opportunity} dataset~\cite{OpportunityDataset} contains human activities recorded in a room with kitchen, deckchair, and outdoors using $72$ sensors on the body and objects. Four subjects were invited to perform five complex and high-level activities, namely, \emph{relaxing (\textbf{RL})}, \emph{early morning (\textbf{EM})}, \emph{coffee time (\textbf{CT})}, \emph{sandwich time (\textbf{ST})} and \emph{cleanup (\textbf{CU})} in this environment. Each subject is recorded with ADL runs and a drill run. The ADL runs consist of temporally unfolding actions without pre-defined rules on how to perform the tasks, and the drill run comprises of a pre-defined set of instructions on how to perform the tasks. We selected ADL runs to verify our model due to its realistic scenarios of human activities. The activity \emph{relaxing} has $40$ samples; for the other four activities, each has $20$ samples. Each activity in our dataset is composed of $30$ low-level actions that have been well-labeled with action $id$, start and end times. In particular, these actions include four body locomotions (\emph{sit, stand, walk, lie}) and $13$ actions for each of the left and right hands (e.g, \emph{close, reach, open, move}). In this experiment, we utilized the $30$ low-level actions as input and the five high-level activities as output. The performance reported in this paper was measured based on 10-fold cross-validation classification accuracy.

\subsection{Performance of Temporal Patterns}
To verify the representativeness of our temporal patterns, we compare the following approaches:
\begin{enumerate}
  \item bag-of-actions: It represents activities as a multiset of actions, discarding the action order but keeping multiplicity.
  \item $1$-$patterns$: Generated by our method.
  \item $\{1,2\}$-$patterns$: Combination of $1$ and $2$-$patterns$.
  \item $\{1,2,3\}$-$patterns$: Combination of $1$, $2$ and $3$-$patterns$.
\end{enumerate}

The results for bag-of-actions are: \textbf{SVM} ($80.0\%$), \textbf{kNN} ($79.8\%$), \textbf{Lasso} ($90.8\%$), \textbf{MTL} ($91.6\%$), \textbf{GL} ($91.5\%$), \textbf{aMTL} ($\textbf{93.3\%}$). The results for temporal patterns are presented in Figure \ref{patternNumberEffects}. From the above results and Figure \ref{patternNumberEffects}, it can be seen that approaches based on $1$-$patterns$ outperform those based on the bag-of-actions approach. This is because $1$-$patterns$ take not only the action frequency into account, but also how long the action appears in this activity. This is intuitive as the longer an action appears, the more important the action will be. In addition, we can observe that the higher-order temporal patterns, such as $\{1,2\}$-$patterns$ and $\{1,2,3\}$-$patterns$ show superiority over others. This demonstrates that the temporal relatedness among the actions is able to enhance the description of activities.

In addition, we performed pairwise significance test among various representation approaches based on the same \textbf{aMTL} model. The results are summarized in Table \ref{t_test_features}. All the $p$-$values$ are smaller than 0.05, which indicates that our proposed temporal pattern mining approach is significantly better than the bag-of-actions, and the improvements by higher patterns are statistically significant.

\begin{table}[t]
\begin{center}
\small
\caption{{Pairwise significance test between different features on \textbf{aMTL}.}}\label{t_test_features}
\begin{tabular}{|c||c|}
  \hline
  Pairwise Significance Test & $p$-$values$ \\
  \hline
  \hline
  $1$-$patterns$ vs bag-of-actions & $9.3e$-$3$ \\
  \hline
  $\{1,2\}$-$patterns$ vs $1$-$patterns$ & $3.5e$-$3$ \\ 
  \hline
  $\{1,2,3\}$-$patterns$ vs $1$-$patterns$ & $3.3e$-$6$ \\
  \hline
\end{tabular}
\end{center}
\end{table}

\begin{table*}[t]
\begin{center}
\small
\caption{{Pairwise significance test between our proposed \textbf{aMTL} model and each of the baselines. For each pair, their $10$-$fold$ results were utilized to conduct the t-test.}}\label{t_test}
\begin{tabular}{|c||c|c|c|c|c|}
  \hline
  Pairwise Significance Test & \textbf{aMTL} vs \textbf{SVM} & \textbf{aMTL} vs \textbf{kNN} & \textbf{aMTL} vs \textbf{Lasso} & \textbf{aMTL} vs \textbf{MTL} & \textbf{aMTL} vs \textbf{GL}  \\
  \hline
  \hline
  $1$-$patterns$ & $1.0e$-$2$ & $1.0e$-$2$ & $4.3e$-$4$ & $2.3e$-$2$ & $1.1e$-$3$ \\ 
  \hline
  $\{1,2\}$-$patterns$ & $2.4e$-$2$ & $9.9e$-$3$ & $7.9e$-$7$ & $2.2e$-$3$ & $3.6e$-$5$ \\
  \hline
  $\{1,2,3\}$-$patterns$ & $3.3e$-$2$ & $8.1e$-$3$ & $1.0e$-$8$ & $2.7e$-$4$ & $2.7e$-$7$ \\

  \hline
\end{tabular}
\end{center}
\end{table*}

\subsection{Learning Model Comparison}
To validate our proposed \textbf{aMTL} model, we compared it with five baselines:
\begin{itemize}
  \item \textbf{SVM}: We implemented this method with the help of LIBSVM\footnote{\small{\url{http://www.csie.ntu.edu.tw/~cjlin/libsvm/}}}. We selected a linear kernel.
  \item \textbf{kNN}: We employed the k-Nearest Neighbors in OpenCV\footnote{\small{\url{http://opencv.org/}}} and set $K = 7$.
  \item \textbf{Lasso}: \textbf{Lasso}~\cite{tibshirani_LASSO1996regression} tries to minimize the objective function $\frac{1}{2} \| \textbf{XW} - \textbf{Y} \| ^{2}_{F} + \alpha\| \textbf{W}\| ^{2}_{F} + \beta\| \textbf{W} \|_{1}$ and encodes the sparsity over all weights in $\textbf{W}$. It keeps task-specific features but ignores the task-sharing features.
  \item \textbf{MTL}: As a typical example of traditional multi-task learning, the work of Zhang and Yeung~\shortcite{ZhangYu_UAI_MTL_2010} aims to capture the task relationship in multi-task learning. We can derive \textbf{MTL} from our model by setting $\theta = 0$.
  \item \textbf{GL}: The last baseline is group Lasso regularization method with a $\ell_{2,1}$-norm penalty for group feature selection $\frac{1}{2} \| \textbf{XW} - \textbf{Y} \| ^{2}_{F} + \mu\| \textbf{W}\| ^{2}_{F} + \nu\| \textbf{W} \|_{2,1}$~\cite{GroupLassoOriginal_Statistic2006}. This model encodes the group sparsity but fails to take task relatedness into account. We can derive \textbf{GL} from \textbf{aMTL} by setting $\lambda = 0$.
\end{itemize}
We retained the same parameter settings over all the experiments. For \textbf{aMTL}, we set $minsup = 0.01$ and $t_{win} = 2\times L_{avg}$ over all the experiments, where $L_{avg}$ is the average length of action intervals in an activity.

\begin{figure}
  \centering
  \includegraphics[width=0.40\textwidth]{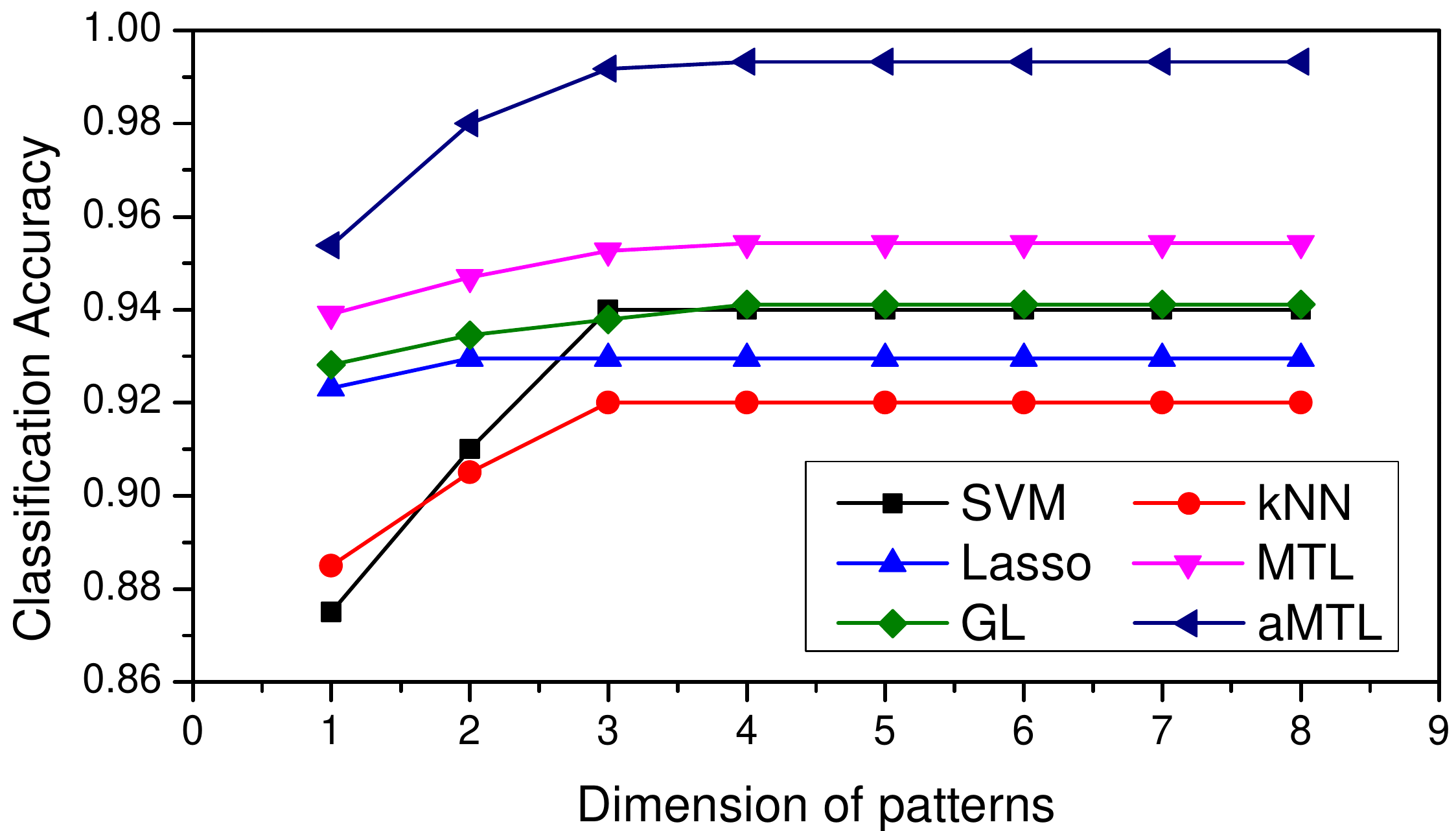}\\
  \caption{Comparative performance illustration of activity recognition with respect to various models on different pattern dimensions.}\label{patternNumberEffects}
\end{figure}

The experimental results are demonstrated in Figure \ref{patternNumberEffects}. From this figure, it can be seen that \textbf{MTL} outperforms the single task learning methods, which verifies that there exists relatedness among these activities and such relatedness can boost the learning performance. Moreover, as compared to the single task learning methods, \textbf{Lasso} achieves a bit higher accuracy due to the fact that the temporal pattern features for representing complex activities are quite sparse. However, as \textbf{Lasso} can only keep the task-specific features, \textbf{GL} shows a slightly better performance. This is because group sparsity is addressed during the learning, and the task-sharing features are also learned. This further justifies the assumption that only a small set of temporal patterns are predictive for activity recognition tasks. In addition, our \textbf{aMTL} model outperforms \textbf{MTL} by $2\%$-$4\%$ in overall accuracy. This is because our model encodes the group sparsity during learning and learns the activity-sharing and activity-specific temporal features simultaneously. Moreover, our method significantly outperforms \textbf{GL} with improvement of $3\%$-$6\%$. This is to be expected since the \textbf{aMTL} model can capture the relatedness among activities and further improve performance. Moreover, the activity relatedness learned by \textbf{aMTL} is illustrated in Table \ref{ActivityRelateness}. The matrix in Table \ref{ActivityRelateness} is encoded in matrix $\bm{\Omega}$ automatically learned by our proposed \textbf{aMTL} model. It has been normalized, and its entries represent the pairwise similarities between activities. Larger value indicates more correlated relations between two activities. The diagonal elements are removed since they are self-correlated and less attractive. From Table \ref{ActivityRelateness}, it can be seen that there exits different relatedness among activities. In particular, \emph{coffee time} has higher correlations with \emph{sandwich time} rather than \emph{early morning}, and \emph{relaxing} is more related to \emph{early morning} rather than \emph{cleanup}. This is consistent with our human perception, which, in turn, further verifies the assumption that there exists relatedness among activities and these relatedness can boost the performance.

\begin{table}[t]
\begin{center}
\small
\caption{Activity relatedness learned by \textbf{aMTL}.}\label{ActivityRelateness}
\begin{tabular}{|c||c|c|c|c|c|}
  \hline
   & \emph{\textbf{RL}} & \emph{\textbf{CT}} & \emph{\textbf{EM}} & \emph{\textbf{CU}} & \emph{\textbf{ST}} \\
  \hline
  \hline
  \emph{\textbf{RL}} & $-$ & $0.101$ & $0.219$ & $0.102$ & $0.106$ \\ 
  \hline
  \emph{\textbf{CT}} & $0.101$ & $-$ & $0.036$ & $0.142$ & $0.212$ \\ 
  \hline
  \emph{\textbf{EM}} & $0.219$ & $0.036$ & $-$ & $0.117$ & $0.008$ \\
  \hline
  \emph{\textbf{CU}} & $0.102$ & $0.142$ & $0.117$ & $-$ & $0.139$ \\
  \hline
  \emph{\textbf{ST}} & $0.106$ & $0.212$ & $0.008$ & $0.139$ & $-$ \\
  \hline
\end{tabular}
\end{center}
\end{table}

We have also performed pairwise significance test between \textbf{aMTL} model and each of the baselines under various dimensions of temporal patterns, and the results are shown in Table \ref{t_test}. It can be seen that all the $p$-$values$ are smaller than 0.05, which demonstrates that our \textbf{aMTL} model is consistently and significantly better than the baselines across various temporal pattern features.

\begin{figure*}
  \centering
  \subfigure[Effects of $\gamma$]{\label{SensitiveGamma}\includegraphics[width=0.30\textwidth]{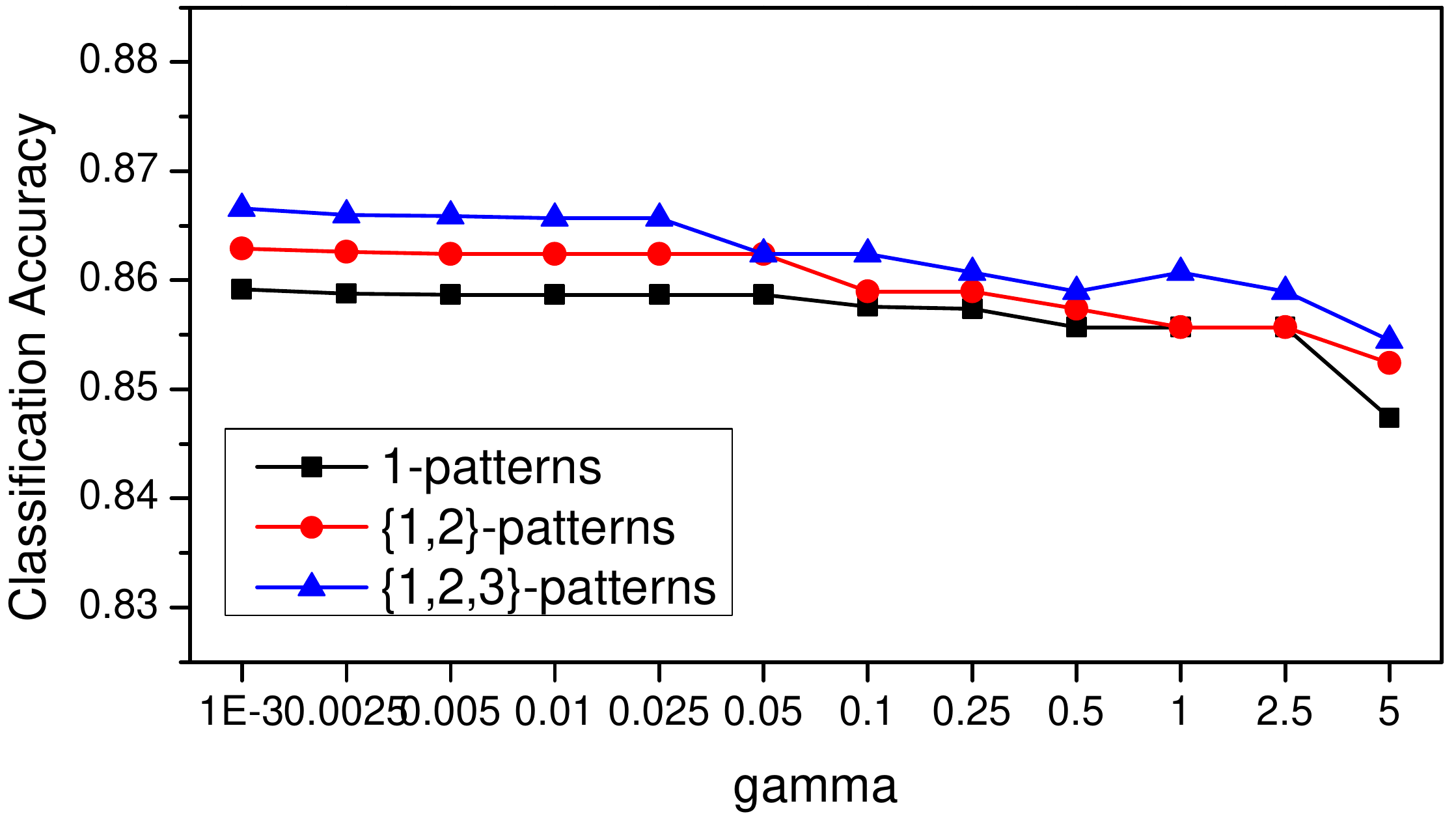}}
  \subfigure[Effects of $\lambda$]{\label{SensitiveLamda}\includegraphics[width=0.30\textwidth]{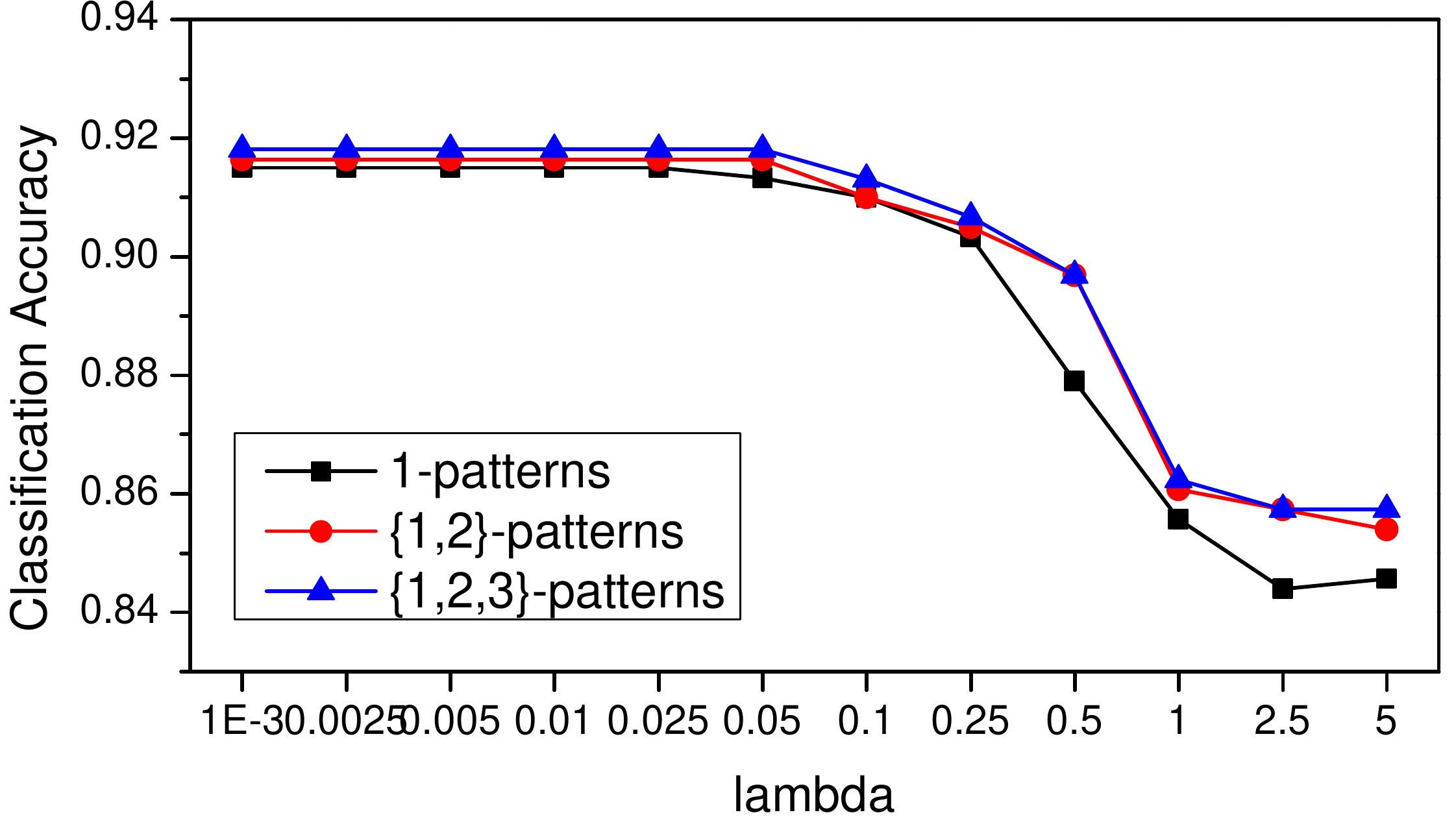}}
  \subfigure[Effects of $\theta$]{\label{SensitiveTheta}\includegraphics[width=0.30\textwidth]{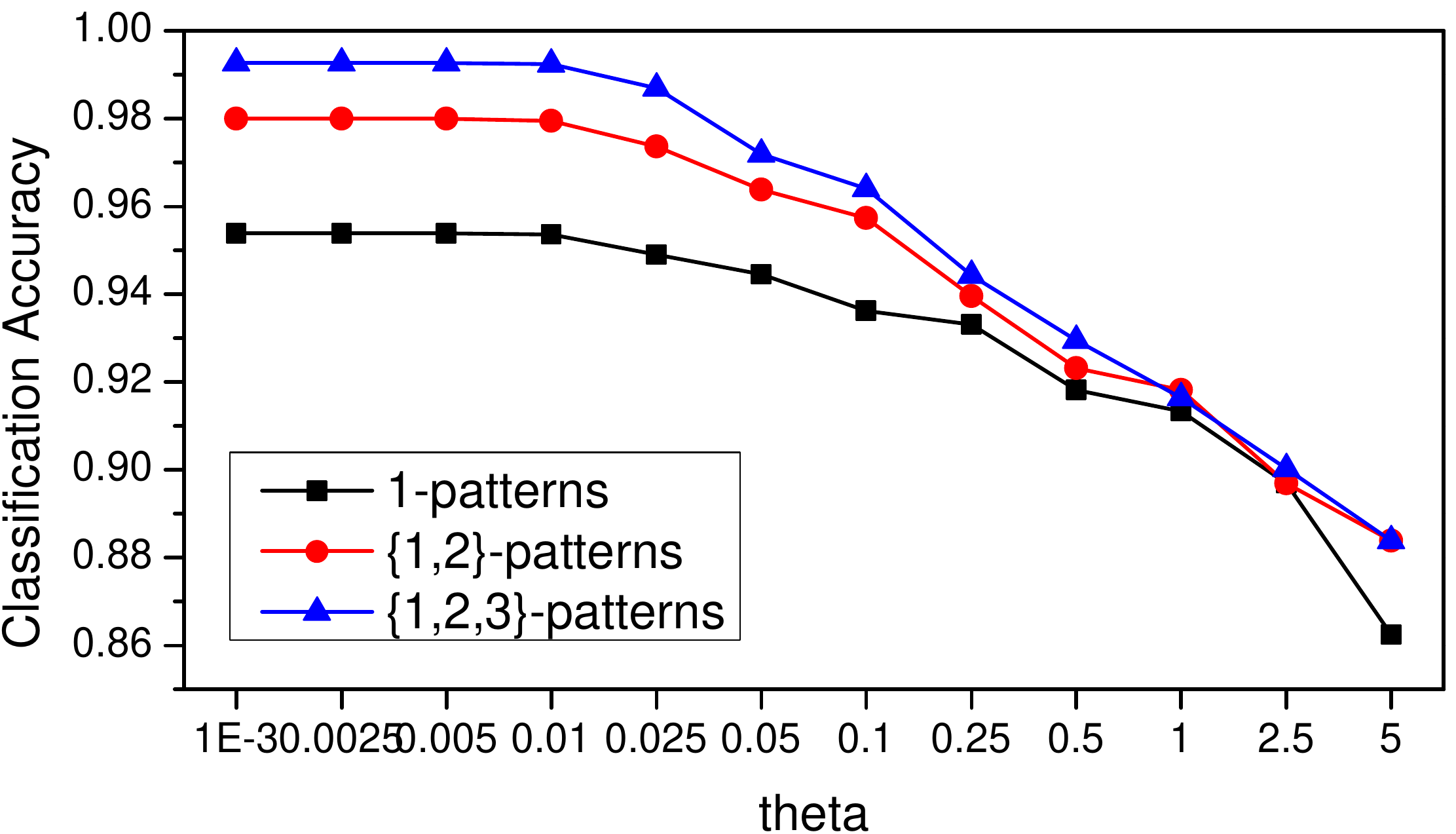}}
  \caption{Performance illustration of activity recognition with respect to the sensitivity of \textbf{aMTL} parameters: $\gamma, \lambda, \theta$.}\label{SensitiveParameters}
\end{figure*}

\subsection{Overall Scheme Evaluation}
To validate our proposed scheme (temporal patterns + \textbf{aMTL}) for activity recognition, we also compared it against three baselines: two dynamical model approaches (\textbf{HMM}~\cite{HMM_rabiner1989tutorial} and \textbf{CRF}~\cite{CRF_lafferty2001conditional}) and a state-of-the-arts activity recognition approach (\textbf{ITBN}~\cite{ITBN_zhang2013modeling}).

The results are displayed in Table \ref{ComparisonWithOtherMethods}. From Table \ref{ComparisonWithOtherMethods}, we have the following observations: 1) \textbf{HMM} performs much worse than other methods due to the following reasons. First, \textbf{HMM} has strong first-order Markovian assumption for the state sequences, which may not be adequate for activity sequences;  Second, it does not consider the complicate pairwise temporal relationships between actions, which may contribute significantly to activity recognition. 2) As compared to \textbf{HMM}, \textbf{CRF} achieves better performance by modeling the pairwise action relations. 3) our approach ($\{1,2\}$-$patterns$ $+$ \textbf{aMTL}) outperforms \textbf{CRF}, since the temporal patterns can capture more complex pairwise relations between actions (e.g., overlapping). This clearly reflects that temporal data contains rich temporal relatedness information than sequential data and this kind of information, in turn, can further boost the performance. Similarly, $\{1,2,3\}$-$patterns$ $+$ \textbf{aMTL} outperforms \textbf{CRF} due to its ability of capturing higher-order temporal relations among actions (i.e., $\{3\}$-$patterns$). And 4) our proposed method achieves higher performance than \textbf{ITBN}. Several reasons lead to this result. First of all, \textbf{ITBN} is a Bayesian network with directed acyclic graph and it will automatically remove samples that contain temporal relation conflicts. This process greatly reduces the training size and hence the training performance. Second, since \textbf{ITBN} tends to remove temporal relation conflicts, it will lose some kinds of temporal relatedness among actions which may affect its performance as well.

\begin{table}[t]
\begin{center}
\small
\caption{Performance comparison with state-of-the-art activity recognition approaches.}\label{ComparisonWithOtherMethods}
\begin{tabular}{|c||c|}
  \hline
  Method & Accuracy $\%$\\
  \hline
  \hline
  \textbf{HMM} & $54\%$ \\
  \hline
  \textbf{CRF} & $95\%$  \\ 
  \hline
  \textbf{ITBN} & $88\%$ \\
  \hline
  $\{1,2\}$-$patterns$ $+$ \textbf{aMTL} & $\textbf{98.0\%}$ \\
  $\{1,2,3\}$-$patterns$ $+$ \textbf{aMTL} & $\textbf{99.2\%}$ \\
  \hline
\end{tabular}
\end{center}
\end{table}

\subsection{Sensitivity of Parameters}
Our algorithm involves several parameters, hence it is necessary to study the performance sensitivity over them.

We first investigated the performance of activity recognition over the pattern dimension. Figure \ref{patternNumberEffects} comparatively illustrates the experimental results of activity recognition with respect to various pattern dimensions. It can be seen that when $k$ is larger than $3$, the accuracy performance tends to be stable. The reason is that as $k$ gets larger, fewer discriminative patterns are found, so the algorithm terminates automatically. Notably, larger patterns usually introduce high-dimensional features and lead to expensive time complexity for feature extraction. These experimental results reveal that $\{1,2,3\}$-$patterns$ are descriptive enough.

We then examined the effects of three key parameters involved in our \textbf{aMTL} model. They are $\gamma, \lambda, \theta$, which respectively balance the trade-off among generalization error, activity-relatedness and group sparsity. We initially fixed $\lambda$ and $\theta$, and then varied $\gamma$ from $0.001$ to $5$ and doubled the value at each step. The experimental results over different $\gamma$ are shown in Figure \ref{SensitiveGamma}. It can be seen that $\gamma$ does not affect performance too much as it only increases a little when $\gamma$ decreases and gives a slightly better result when $\gamma = 0.001$. We then set $\gamma = 0.001, \theta = 1$ and varied $\lambda$. The results are illustrated in Figure \ref{SensitiveLamda}. It can be seen that the performance increases as $\lambda$ decreases but stabilizes at $0.05$. Finally, we set $\gamma = 0.001, \lambda = 0.05$ and varied $\theta$. We can see that small $\theta$ leads to better performance, as illustrated in Figure \ref{SensitiveTheta}. However, it remains steady when $\theta$ equals or is less than $0.01$.

\subsection{Discussion}
In this section, we discuss the efficiency and scalability of our proposed scheme. The efficiency of temporal pattern mining algorithm is guaranteed by its convergence at $\{1,2,3\}$-$patterns$ as shown in Figure \ref{patternNumberEffects}, i.e., we can stop at $\{3\}$-$patterns$. On the other hand, the computational cost of \textbf{aMTL} is not expensive. In particular, for the optimization of Eqn.$(\ref{OptimizingW})$, the complexity for each iteration in the FISTA algorithm is $O((N+M)DM)$. Moreover, the FISTA algorithm converges within $O(1/\epsilon^{2})$ iterations, where $\epsilon$ is the desired accuracy. For the optimization of Eqn.$(\ref{OptimizingOmega})$, calculating the closed-form solution in Eqn.$(\ref{OmegaAnalyticaSolution})$ scales up to $O(DM^{2}+M^{3})$. Therefore, the total time cost for one iteration in the alternating algorithm is $O(\frac{(N+M)DM}{\epsilon^2} + DM^{2} + M^{3})$. From the experiments, we find that the alternating algorithm only needs very few iterations to converge (usually within 20 iterations), making the whole procedure very efficient.

Our proposed scheme is also scalable to big data and other activity recognition. The scalability of temporal pattern mining algorithm is ensured by its pruning step, since it tends to select only frequent temporal patterns from the activities, which makes the algorithm scalable to large datasets. For the \textbf{aMTL} algorithm, according to its time complexity analysis, it is linearly dependent to the number of features $D$ and has a relation $O(M^3)$ with the number of activities. In most of the real applications, we only focus on a small number of important activities. Therefore, our proposed scheme can easily scale to large datasets with high-dimensional features. In addition, the relatedness among different activities are automatically learned form given dataset. Hence our scheme is generalizable to other activity recognition.

\section{Conclusion and Future Work}
This paper presents a scheme to recognize activities from sensor data. It comprises of two components. The first component is temporal pattern mining. It works towards mining frequent patterns from low-level actions. The second one trains an adaptive multi-task learning model to capture the relatedness among activities. Extensive experiments on real-world data show significant gains of these two components and their overall performance as compared to state-of-the-arts methods.

In future, we plan to extend our work to deal with error propagation problems in the output of low-level action recognition systems.

\section*{Acknowledgments}
This work was supported in part by grants R-252-000-473-133 and R-252-000-473-750 from the National University of Singapore.

\small
\bibliographystyle{named}
\bibliography{ijcai15}
\end{document}